\documentclass{ifacconf}
\usepackage{xcolor}
\usepackage{graphicx}      
\usepackage{natbib}        
\usepackage{import}        
\usepackage{amsmath,amsfonts}
\usepackage{comment}
\usepackage{url}

\newcommand{\ba}{{\mathbf a}}
\newcommand{\bb}{{\mathbf b}}
\newcommand{\bx}{{\mathbf x}}
\newcommand{\bX}{{\mathbf X}}
\newcommand{\bA}{{\mathbf A}}

\newcommand{\bU}{{\mathbf U}}
\newcommand{\bG}{{\mathbf G}}

\newcommand{\bv}{{\mathbf v}}

\newcommand{\bF}{{\mathbf F}}

\newcommand{\bp}{{\mathbf p}}

\newcommand{\bK}{{\mathbf K}}

\newcommand{\bM}{{\mathbf M}}

\newcommand{\by}{{\mathbf y}}
\newcommand{\bz}{{\mathbf z}}
\newcommand{\bZ}{{\mathbf Z}}
\newcommand{\bh}{{\mathbf h}}

\newcommand{\bY}{{\mathbf Y}}
\newcommand{\bg}{{\mathbf g}}
\newcommand{\bB}{{\mathbf B}}
\newcommand{\bu}{{\mathbf u}}

\newcommand{\bR}{{\mathbf R}}

\newcommand{\bJ}{{\mathbf J}}
\newcommand{\bH}{{\mathbf H}}

\newcommand{\bff}{{\mathbf f}}

\newcommand{\mC}{{\mathbb C}}
\newcommand{\mR}{{\mathbb R}}

\newcommand{\cK}{{\cal K}}

\newcommand{\cU}{{\cal U}}

\newcommand{\cX}{{\cal X}}

\newcommand{\cN}{{\cal N}}

\newcommand{\bPsi}{{\boldsymbol \Psi}}
\newcommand{\bC}{{\boldsymbol{\mathcal C}}}

\newcommand{\bomega}{{\boldsymbol \omega}}

\DeclareMathOperator*{\argmin}{arg\,min}

\begin{document}
\begin{frontmatter}

\title{SE(3) Koopman-MPC: Data-driven Learning and Control of Quadrotor UAVs}


\author[First]{Sriram S. K. S. Narayanan}, 
\author[First]{Duvan Tellez-Castro}, 
\author[First]{Sarang Sutavani},
\author[First]{Umesh Vaidya}

\address[First]{Mechanical Engineering Department, Clemson University, 
   Clemson, SC 29634 USA \\
   (e-mail: \{sriramk, dtellez, ssutava, uvaidya\}@clemson.edu).}

\begin{abstract}                
In this paper, we propose a novel data-driven approach for learning and control of quadrotor UAVs based on the Koopman operator and extended dynamic mode decomposition (EDMD). Building observables for EDMD based on conventional methods like Euler angles (to represent orientation) is known to involve singularities. To address this issue, we employ a set of physics-informed observables based on the underlying topology of the nonlinear system. We use rotation matrices to directly represent the orientation dynamics and obtain a lifted linear representation of the nonlinear quadrotor dynamics in the SE(3) manifold. This EDMD model leads to accurate prediction and can be generalized to several validation sets. Further, we design a linear model predictive controller (MPC) based on the proposed EDMD model to track agile reference trajectories. Simulation results show that the proposed MPC controller can run as fast as 100 Hz and is able to track arbitrary reference trajectories with good accuracy. Implementation details can be found in \url{https://github.com/sriram-2502/KoopmanMPC_Quadrotor}.
\end{abstract}


\end{frontmatter}

\section{Introduction}\label{Introduction}
Designing optimal controllers for nonlinear robotic systems is a challenging problem. The goal is to develop a controller that can stabilize the system and follow reference trajectories while adhering to constraints arising due to system dynamics and actuator limits. Model predictive control (MPC) is an optimization-based method that can be used to achieve this goal (\cite{rawlings2017model, berberich2022linear}). The ability to specify performance metrics and constraints makes it very intuitive when designing controllers for nonlinear systems. Although recent advances in optimization techniques and computing power proposed by \cite{kouzoupis2018recent, gros2020linear} have made it possible to implement nonlinear model predictive controllers (NMPC) in real-time, they still require sufficiently accurate mathematical models and cannot handle model uncertainty. To address some of these issues, \cite{hou2013model, hewing2019cautious, 
 krolicki2022nonlinear} have developed data-driven techniques, and \cite{lusch2018deep, han2020deep, wang2021deep} have developed learning-based methods to identify the underlying model of the system. However, these methods are computationally intensive and cannot be applied to general high-dimensional nonlinear systems. 

To overcome the issues associated with NMPC, we propose a Koopman operator based approach to learn linear predictors for a controlled nonlinear dynamical system. The Koopman operator governs the evolution of a set of observable functions, which can be interpreted as nonlinear measurement functions of the system states. {\color{black} This} results in a linear (but possibly infinite-dimensional) representation of the underlying nonlinear system (\cite{lan2013linearization, mauroy2016linear}). In \cite{schmid2010dynamic}, the dynamic mode decomposition (DMD), a data-driven approach to obtain a finite-dimensional approximation of the Koopman operator, was proposed. This method uses time-shifted snapshots (measurements) of the system states to approximate the Koopman operator in a least-square fashion. This method can be limiting and sometimes fail to capture all the nonlinearities of the system. \cite{williams2015data} proposed extended DMD (EDMD) in which snapshots of nonlinear measurement functions (observables) can be augmented with the system states to obtain a ``lifted" finite-dimensional approximation of the Koopman operator. Recently, \cite{korda2018linear} extended EDMD for controlled dynamical systems; \cite{sinha2019computation} proposed a framework to compute Koopman operators from sparse data. One of the main challenges in developing EDMD-based approximation methods is the choice of observable functions that can capture the underlying nonlinear dynamics. Recent works from \cite{chen2022koopman,zinage2021koopman,zinage2022koopman} have proposed the use of physics-informed observables for a quadrotor system that preserves the underlying topology. However, they still rely on accurate knowledge of the model and hence cannot handle model uncertainties.

In this work, we propose an EDMD-based approach to identify the underlying system dynamics of a quadrotor using physics-informed observable functions and develop a linear MPC to track reference trajectories. Recent works have successfully developed EDMD-based methods to obtain linear predictors for nonlinear robotic systems. \cite{bruder2019nonlinear} developed an EDMD-based system identification method for a soft robotic arm; \cite{bruder2020data} used a linear MPC for control. Other works from \cite{shi2021acd} used an analytical construction of observables based on Hermite polynomials for a Dubin's car model and a two-link robotic arm; \cite{shi2022online} extended it to control soft multi-fingered grippers using MPC. However, these methods are hard to extend to quadrotors since their underlying topology is quite distinct. \cite{folkestad2021koopman, folkestad2022koopnet} proposed a learning-based approach for approximating the Koopman operator and NMPC to track reference trajectories. \cite{mamakoukas2022robust} proposed a robust MPC to improve constraint satisfaction in the presence of uncertainties.  However, these methods are slow and difficult to scale for agile quadrotors.

To this extent, we propose a linear MPC-based controller for the agile locomotion of quadrotor systems. We identify the Koopman linear predictors from data using physics-informed observable functions. The physics-informed observables are constructed using the product of rotation matrices and the angular velocities of the quadrotor system. We treat the angular orientation of the quadrotor using a rotation matrix which is an element of the Lie group SO(3). Further, we augment the linear states to this set of observables, ensuring that the "lifted" states have the same underlying topology as the original system. We then use this set of observables to learn the linear predictors of the quadrotor system using EDMD. Further, we design a linear MPC in the lifted space using these linear predictors to track arbitrary reference trajectories.

\subsection{Main Contribution}
The main contribution of this paper is the data-driven identification of the quadrotor dynamics based on EDMD and the design of a linear MPC to track reference trajectories. We use EDMD to obtain the Koopman linear predictors of the controlled nonlinear dynamical systems using physics-informed observable functions. These observables are singularity-free and preserve the system's underlying SE(3) topology. Further, we design a linear MPC in the lifted space, which can be implemented in real time using a quadratic program. Finally, we use this proposed framework to track arbitrary reference trajectories.

\subsection{Organization}
The rest of the paper is organized as follows. In Section \ref{preliminary}, we introduce the preliminaries used in this paper. Section \ref{sec:quad} outlines the dynamics of the quadrotors. Section \ref{sec:koopman} presents the basics of Koopman operator theory, followed by its finite-dimensional approximation in Section \ref{sec:EDMD}. Section \ref{sec:MPC} formulates a linear MPC in the "lifted space," which can be solved as a Quadratic Program, as mentioned in Section  \ref{sec:qp}. In Section \ref{results}, we learn the Koopman linear predictors for the quadrotor system and show MPC results for tracking random reference trajectories. Section \ref{sec:EDMD_results} describes the training and validation procedures to learn the Koopman linear predictors from data; Section \ref{sec:MPC_results} shows the performance of the MPC designed using the learned linear predictors for tracking random reference trajectories. Finally, we conclude the results of this work in Section \ref{conclusions}.

\section{Preliminaries} \label{preliminary}

In this section, we introduce the dynamics of the quadrotor, the theory behind the Koopman operator and its finite-dimensional approximation for nonlinear systems with control using the EDMD algorithm. 

\subsection{Quadrotor Dynamics} \label{sec:quad}
  \begin{figure}
     \centering
     \includegraphics[width = 0.6\linewidth]{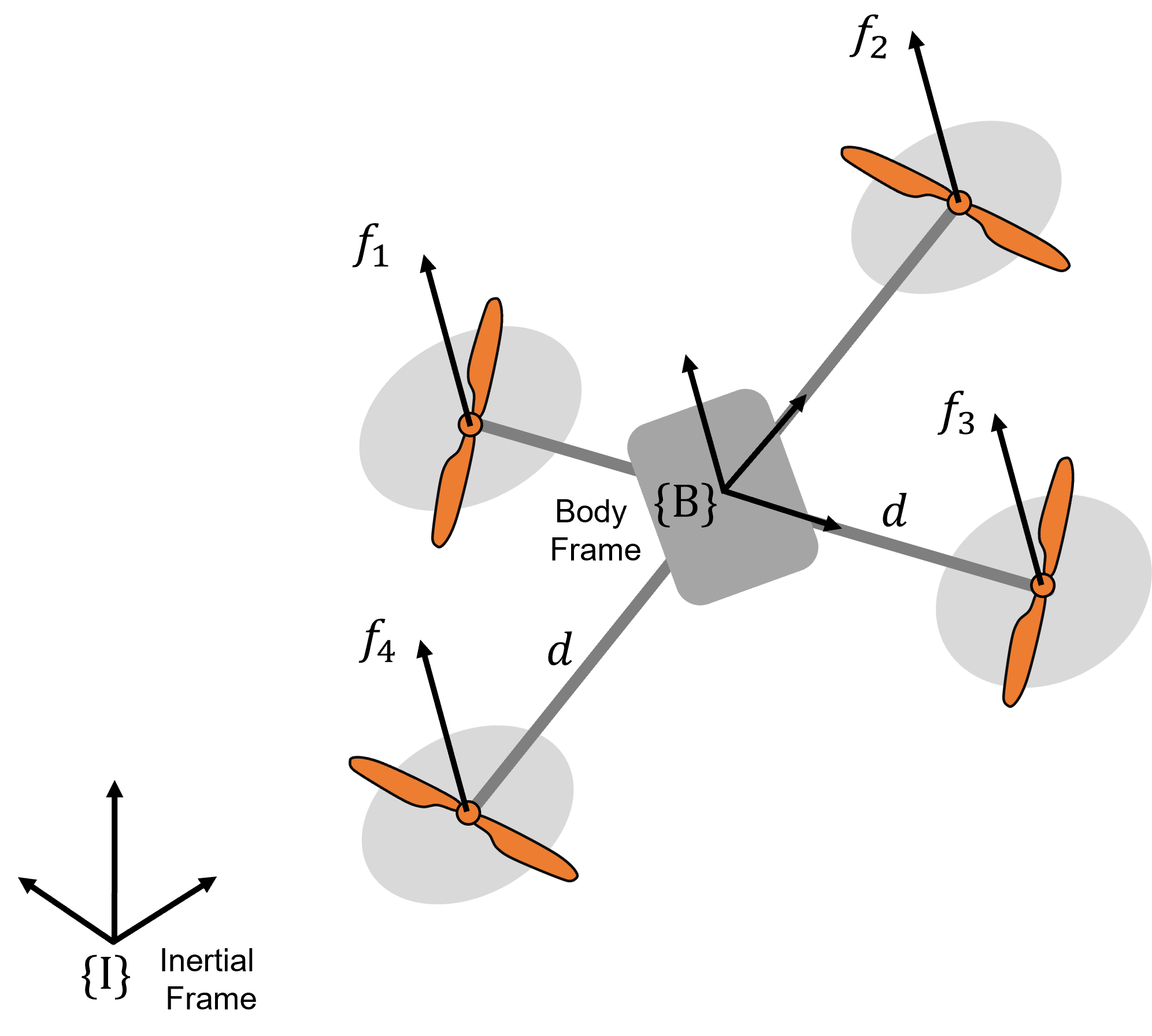}
     \caption{Quadrotor system.}
     \label{fig:schematic}
 \end{figure}
 
Consider the quadrotor system (shown in Fig. \ref{fig:schematic}) composed of linear and attitude dynamics. The linear dynamics of the system can be described by the states $[\bp,\;\bv]$ represented in the {\color{black}inertial} frame, where $\bp \in \mathbb{R}^3$ represents the position states and $\bv \in \mathbb{R}^3$ represents the velocity states. The net thrust generated by the propellers in the z-direction $f_t = \sum_i f_i \in \mathbb{R}$, represented in the body frame, is given as the input to the linear dynamics. Note that the angular dynamics of the system can be described using the states $[{\color{black}\bR},\;{\bomega}]$ where ${\color{black}\bR} \in \mathbb{R}^{3 \times 3}$ is a rotation matrix that maps from the body-fixed frame to the inertial frame, and $\bomega \in \mathbb{R}^3$ is the angular velocity represented in the body-fixed frame. Note that $R$ is an element of the special orthogonal group SO(3) such that $R^\top {\color{black}\bR} = I_{3 \times 3}$ (where $I$ is the identity matrix) and $\det({\color{black}\bR})=1$. The control input to the attitude dynamics is the total moment ${\color{black}\bM} \in \mathbb{R}^3$, represented in the body-fixed frame. The overall dynamics of the system are given by
\begin{align}
    &\dot{\bp} = \bv \nonumber \\ 
    &{\color{black} \dot{\bv} = \frac{1}{m}\bff_t \bR - \bg} \nonumber \\
    &\dot{{\color{black}\bR}} = {\color{black}\bR}\hat{\bomega} \nonumber \\
    &\dot{\bomega} = {\color{black}\bJ}^{-1}({\color{black}\bM} - \bomega \times {\color{black}\bJ}\bomega), \label{eqn:quad_eom}
\end{align}
where $m \in \mathbb{R}$ is the mass of the quadrotor, ${\color{black}\bJ} \in \mathbb{R}^{3 \times 3}$ is the inertia represented in the body-fixed frame, $\bff_t = [0,\;0,\;f_t]^\top$ and $\bg = [0,\;0,\;g]^\top$. The operator $\hat{(\cdot)}:\mathbb{R}^3 \xrightarrow{}\text{SO}(3)$ represents the hat map defined on the cross product of two vectors ${\color{black}\ba} \times {\color{black}\bb} = {\color{black}\hat{\ba}\bb}$ for ${\color{black}\ba,\bb} \in \mathbb{R}^3$. 

{\color{black}We assume the that torque generated} by each propeller $\tau_i$ is directly proportional to its thrust $f_i$. Further, we assume that propellers 1 and 3 rotate clockwise and propellers 2 and 4 rotate counterclockwise. Hence the thrust produced by each proper can be obtained as $f_i = (-1)^i c_{\tau}$, where $c_{\tau} \in \mathbb{R}$ is a scalar constant. Further, the control input for the linear and attitude dynamics ($f_t, {\color{black}\bM}$) can be obtained from the thrust produced by each propeller as follows
\begin{equation}
    \begin{bmatrix}
        f_t \\ M_1 \\ M_2 \\ M_3
    \end{bmatrix} =
    \begin{bmatrix}
        1 & 1 & 1 & 1 \\
        0 & -d & 0 & d \\
        d & 0 & -d & 0 \\
        -c_{\tau} & c_{\tau} & -c_{\tau} & c_{\tau}
    \end{bmatrix}
    \begin{bmatrix}
        f_1 \\ f_2 \\ f_3 \\ f_4
    \end{bmatrix},
\end{equation}
where $d \in \mathbb{R}$ is the distance between the center of mass of the quadrotor and the center of each propeller. The physical parameters of the quadrotor used in this paper are listed in Table \ref{tab:quad_params}.

 \begin{table}[ht]
\caption{Physical parameters for the quadrotor}
\centering
\begin{tabular}{ c c c c } 
 \hline
 Parameter & Variable & Value & Units \\
\hline
 mass           & $m$        & 4.34   & kg \\ 
 inertia        & $I_{xx}$   & 0.0820 & kg-m$^2$ \\ 
 inertia        & $I_{yy}$   & 0.0845 & kg-m$^2$ \\ 
 inertia        & $I_{zz}$   & 0.1377 & kg-m$^2$ \\
 moment arm     & $d$        & 0.315  & m \\
 torque coeff.  & $c_{\tau}$ & 8$\times 10^{-4}$  & {\color{black}m} \\
 gravity        & $g$        & 9.81  & m/s$^2$ \\
 \hline
\end{tabular}
\label{tab:quad_params}
\end{table}

\subsection{Koopman Operator Theory} \label{sec:koopman}
The Koopman operator and its spectral properties have a wide variety of applications in dynamical systems and control (\cite{lan2013linearization, huang2020data, huang2018feedback, vaidya2022spectral}). Consider a nonlinear system $\bx_{t+1} = \bff(\bx_{t})$
where $\bx \in \cX \subseteq \mR^n$ and $\bu \in \cU \subseteq \mR^m$ ; $\bff : \cX \rightarrow \cX$ defines the vector field which propagates the system states. For such a dynamical system, there exists a set of nonlinear observable functions $\psi(x)$, along which the system evolves linearly. This evaluation is governed by an infinite-dimensional operator $\cK$, referred to as the Koopman operator
\begin{gather}
    [\cK \psi](x) = \psi \circ\bff(x). \label{eqn:Koopman_Operator}
\end{gather}
The Koopman operator can be easily extended to systems with control, $\bx_{t+1} = {\color{black}\bF}(\bx_t, \bu_t)$, {\color{black}where $\bF: \cX \times \cU \rightarrow \cX$}, by defining an augmented state vector $ \bar \bx = [\bx^\top, \bu^\top]^\top$. 
In the following section, we describe the procedure to obtain a finite-dimensional approximation for $\cK$ using the EDMD approach.

\subsection{EDMD with control} \label{sec:EDMD}
 We use the EDMD method (\cite{korda2018linear, proctor2016dynamic}) to obtain an approximation $\bK$ of the infinite-dimensional Koopman operator $\cK$ by projecting it onto a finite subspace of observable functions. This can be achieved using a least square regression. 
 The linear predictor $\bK$ is composed of $\bA \in \mR^{N \times N}$ and $\bB \in \mR^{N \times m}$ which can be used to describe the linear evolution of the observable functions
 \begin{align}\label{eqn:Lifted_linear_dynamics_psi}
      &\bPsi(\bx_{t+1}) = \bA \bPsi(\bx_t) + \bB \bu_t \nonumber \\
      &\bPsi(\bx_{t+1}) = \begin{bmatrix}\bA & \bB \end{bmatrix} \begin{bmatrix}
          \bPsi(\bx_t) \\ \bu_t
      \end{bmatrix} = \bK \bPsi_a(\bx_t, \bu_t),
\end{align}
where $\bPsi(\bx_t)$ = $[\psi_{1}(\bx_t), \ldots, \psi_{N}(\bx_t)]^\top \in \mR^{N}$ ($N>>n$) is the set of observable functions $\psi_i: \cX \rightarrow \mC$. These functions represent a finite-dimensional basis in the lifted space. Using snapshots of the system states and control inputs, we obtain a finite-dimensional approximation of $\bK$ for a controlled nonlinear system. Let $\bX$ and $\bY$ be the pair of datasets generated by the vector field ${\color{black}\bF}$, i.e., $\bX = [\bx_1, \cdots, \bx_M]$, $\bY = [\by_1, \cdots, \by_M]$
such that $\bx_i,\by_i \in \cX$ and $\by_i = \bF(\bx_i,\bu_i)$, {\color{black} where $\bu_i\in \cU$ is some predefined control input}.  
A finite-dimensional approximation of the Koopman operator is then obtained as the minimizer of the following optimization problem.
\begin{align}
    \bK 
    &= \argmin_{ \bK} \| \bPsi(\bY) -  \bK {\bPsi}_a(\bX) \|_{\bF}^{{\color{black}{2}}},
    \nonumber \\
    &=
    \argmin_{ \bK} \sum_{i=1}^{M} \|\bPsi(\by_i) -  \bK {\bPsi}_a(\bx_i)\|_{2}^{2} . \nonumber
\end{align}
By constructing $\bG_1$ and $\bG_2$ such that 
\begin{gather}
    \bG_1 = \frac{1}{M} \sum_{i=1}^{M} \bPsi(\by_i) {\bPsi}_a(\bx_i)^\top, \nonumber
    \\
    \bG_2 = \frac{1}{M} \sum_{i=1}^{M} {\color{black}\bPsi_a(\bx_i)}{\bPsi}_a(\bx_i)^\top. \nonumber
\end{gather}
The solution can be obtained analytically as,
\begin{gather}
    \bK = \bG_1 \bG_2^{\dagger} \nonumber
\end{gather}
{\color{black}where $\bG^{\dagger}$ defines the pseudo-inverse of $\bG \in \mathbb{R}^{m \times n}$.}
\subsection{Linear MPC for Trajectory Tracking} \label{sec:MPC}
We design a linear MPC  in the lifted states for tracking reference trajectories using a linear predictor. Consider the discrete-time system defined by the Koopman linear predictors
\begin{subequations}
\begin{align}
    &{\bz}_{k+1} = \bA {\bz}_{k} + \bB {\bu}_{k} \\
    &{\bx}_{k} = \bC {\bz}_{k}
    \label{eqn:lifted_dynamics}
\end{align}
\end{subequations}
where $\bx_k$ are the original states and {\color{black}$\bz_k = \bPsi(\bx_k)$ are the lifted states}. {\color{black}$\bC \in \mathbb{R}^{N \times n}$} matrix maps the lifted states back to the original states. To track reference trajectories ${\bx}_{ref}$, we design an optimal sequence of controllers that minimize a cost function in a receding horizon fashion.  At time-step $k$, we lift the original states $\bx_k$ to get $\bz_k$ using the observable functions $\Psi(\cdot)$ and the optimal control $\bu_k$ can be obtained by solving the following optimization problem.
\begin{subequations}
\label{eqn:opt}
    \begin{align}
    {\color{black}\underset{u_k, z_k}{\min} \sum_{k=1}^{N_h} \biggl(\tilde{\bz}_k^\top {\bar{Q}} \tilde{\bz}_k + \bu_k^\top \bar{R} \bu_k\biggr)},
    \label{eqn:cost}
    \end{align}
    subject to
    \begin{align}
    &{\bz}_{k+1} = \bA {\bz}_{k} + \bB {\bu}_{k} \label{eqn:MPC_equality}, \\
    & A_{ineq}\bu_k \leq b_{ineq} \label{eqn:MPC_inequality}, \\
    &\bz_0 = {\color{black}\bPsi(\bx_0)} \label{eqn:MPC_ic}.
    \end{align}
\end{subequations}
Here, $\tilde{\bz}_k = \bz_k - \bz_{k_{ref}}$, $\bar{Q} \in \mathbb{R}^{N \times N}$ represents the lifted state cost matrix given by $[Q \quad \mathbf{0}; \mathbf{0} \quad \mathbf{0}]$ with $Q \in \mathbb{R}^{n \times n}$ penalizing the system states and $\bar{R} \in \mathbb{R}^{m \times m}$ is the control cost matrix.
{\color{black}\eqref{eqn:MPC_equality} is the dynamic constraint, \eqref{eqn:MPC_inequality} is the actuator constraints and \eqref{eqn:MPC_ic} is the initial condition constraint.}
\subsection{QP Formulation} \label{sec:qp}
Since the optimization problem has a quadratic cost with linear constraints, the solution can be obtained using a quadratic program with only $\bu_k$ as the decision variable. Hence the dynamics constraint in \eqref{eqn:lifted_dynamics} can be integrated with the cost function as follows
\begin{subequations}
\begin{align}
    {\color{black} J(\bU)} &= {\color{black}\tilde{\bZ}^\top\mathbf{{\bar{Q}}} \tilde{\bZ} + \bU^\top \mathbf{{\bar{R}}} \bU} \\
    \bZ &= \bA_{qp}\bz_0 + \bB_{qp}\bU \\
    \bA_{qp} &= \begin{bmatrix}
    \bA & \bA^2 & \hdots{} & \bA^{N_h} 
    \end{bmatrix} ^\top \nonumber \\
    \bB_{qp} &= \begin{bmatrix}
        \bB & \mathbf{0} &\hdots &\mathbf{0} \\
        \bA\bB & \bB & \hdots &\mathbf{0}\\
        \vdots{} & \vdots{} & \ddots & \vdots{} \\
        \bA^{N_h-1}\bB & \bA^{N_h-2}\bB & \hdots &\bB \nonumber
    \end{bmatrix}
\end{align}
\end{subequations}
where $\mathbf{\bar{Q}}$ is a block diagonal matrix of weights for state deviations $\bar{Q}$, $\mathbf{\bar{R}}$ is a block diagonal matrix of weights for control magnitude $\bar{R}$. {\color{black}$\tilde\bZ$, $\bZ$}, and $\bU$ are stacked vectors of state and control inputs {\color{black}$\tilde{\bz}_k$}, $\bz_k$ and $\bu_k$ respectively over the prediction horizon $N_h$. $\bA_{qp}$ and $\bB_{qp}$ are the stacked matrices for state and control inputs. Hence, the optimization problem can be written in a standard QP form as follows
\begin{subequations}
\begin{align}
    \min_{\bU} \hspace{5mm} &\frac{1}{2}\bU^\top \bH \bU + \bU^\top \bG \\
    \textrm{subject to} \hspace{5mm} & \bA_{ineq} \bU \leq \mathbf{b}_{ineq} 
\end{align}
\end{subequations}
where $\bA_{ineq}$ and $\mathbf{b}_{ineq}$ are stacked block diagonal matrices of constraints, $\bH$ and $\bG$ are the QP matrices given by
\begin{subequations}
    \begin{align}
        &\bH = 2(\bB_{qp}^\top \mathbf{{\bar{Q}}} \bB_{qp} + \mathbf{\bar{R}}) \\
        &\bG = 2\bB_{qp}^\top \mathbf{{\bar{Q}}} (\bA_{qp}\bz_0 - \by)
    \end{align}
\end{subequations}
where $\bA_{qp}$ and $\bB_{qp}$ are the stacked matrices and $\by$ is a stacked vector obtained by lifting the reference trajectory {\color{black}$\bz_{ref}$} for horizon length $N_h$. The optimal control $\bu$ obtained from this optimization at each time-step $k$ is used to propagate the original nonlinear system.

\section{Linear Predictors for quadrotors} \label{results}
In this section, we describe the selection of physics-informed observable functions for the quadrotor dynamics and perform EDMD to obtain linear predictors $\bA$ and $\bB$ for a nonlinear quadrotor system. We evaluate the predictions on trajectories generated using random control inputs on both training and validation sets. We then develop an MPC framework based on the learned linear predictors to track arbitrary reference trajectories.

\subsection{EDMD on SE(3)} \label{sec:EDMD_results}
Based on the quadrotor dynamics defined in equation \eqref{eqn:quad_eom}, it is easy to note that the linear dynamics behave like a double integrator system with no nonlinearities involved. To get a lifted linear mapping for the nonlinear attitude dynamics, inspired by \cite{chen2022koopman}, we build a set of $p$ observable functions ${\color{black}\bar{\bPsi}(\bx)} = [\underline{\bf h}_1^\top,\; \underline{\bf h}_2^\top, \hdots, \underline{\bf h}_p^\top]$ where ${\color{black}\bh_i} \in \mathbb{R}^{3 \times 3}$ {\color{black}with elements} ${\color{black}\bh_1 = \bR\hat{\bomega}, \; \bh_2 = \bR\hat{\bomega}^2, \; \hdots, \bh_p = \bR \hat{\bomega}^p}$. {\color{black}The operator $\underline{(\cdot)}:\mathbb{R}^{m\times m} \xrightarrow{}\mathbb{R}^{m^2 \times 1}$ maps a matrix to a vector by stacking columns of the matrix}. Hence, $\underline{\bf h}_i$ is the vector obtained by stacking columns on $\bh_i$. We augment the original system states along with $\bar{\Psi}(\bx)$ to obtain the final set of lifted states for EDMD
\begin{equation}
    \bPsi(\bx) = [\bp,\; \bv, \; \underline{\bR}^\top, \; \underline{\hat{\bomega}}^\top, \; {\color{black}\bar{\bPsi}(\bx)}]^\top \in \mathbb{R}^{n + 9p}, \nonumber
\end{equation}
where $n=24$. These set of observable functions, when augmented with the linear states, preserve the underlying geometry of SE(3) and hence lead to accurate predictions using the linear predictors $\bA$ and $\bB$ obtained using EDMD. Further, we can retrieve the original states back from the lifted states using {\color{black}$\bC = [I_{n \times n} \quad \mathbf{0}_{n \times 9p}]$}. The angular orientation $\mathbf{\Theta}$ and angular velocity $\bomega$ can be recovered from the states ${\color{black}\bR}$ and $\hat{\bomega}$ 
\begin{subequations}
\begin{align}
    &\mathbf{\Theta} = {\color{black}\log(\bR)^\vee}, \\
    &\bomega = \hat{\bomega}^\vee,
\end{align}
where the operator $(\cdot)^\vee :  \text{SO}(3) \xrightarrow{} \mathbb{R}^3$ is the inverse hat map (also known as the vee map) such that $\hat{a}^\vee = a$.
\end{subequations}

\begin{figure}[htbp]
     \centering
     \includegraphics[width = 1\linewidth]{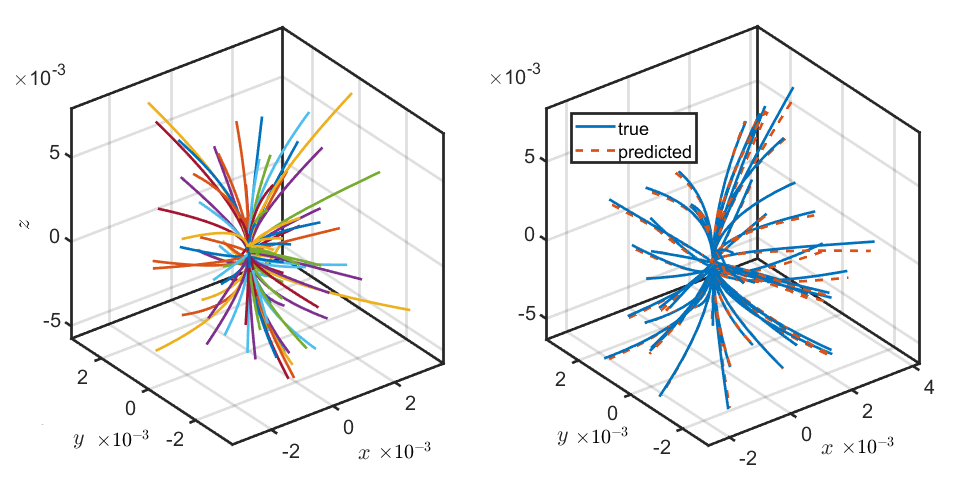}
     \caption{(a) Random trajectories, which represent the training dataset for EDMD, (b) Predicted trajectories using the learned linear Koopman predictors over 100 timesteps on the validation set.}
     \label{fig:EDMD}
 \end{figure}

To learn the linear predictors $\bA$ and $\bB$ through EDMD, we obtain a training dataset by simulating 100 trajectories for $t = 0.1$ s with a sampling rate of $t_s = 0.001$ s as shown in Fig. \ref{fig:EDMD}a. Each trajectory is obtained by forward simulating the quadrotor dynamics (defined in equation \eqref{eqn:quad_eom}) with random control inputs sampled from a normal distribution $[f_t, {\color{black}\bM}]^\top\sim \cN\bigl([0;0;0;0]^\top, \text{diag}([10;10;10;10])\bigr)$. The mean and covariance are selected such that the net thrust $f_t > mg$ for liftoff and the control inputs do not violate the lower and upper bounds on the control inputs. This generates stable flight trajectories for the quadrotor for the specified duration. To evaluate the performance of the learned predictors, we calculate the {\color{black} normalized} Root Mean Square Error (nRMSE)
\begin{equation}
    \text{nRMSE} = 100\times \frac{\sqrt{\sum_k ||\bx_{pred}(kt_s)-\bx_{true}(kt_s)||_2^2}}{\sqrt{\sum_k ||\bx_{true}(kt_s)||_2^2}}. \nonumber
\end{equation}

To learn the Koopman predictors, we use $p=3$ basis functions and augment the control to obtain the extended states $\bar\bx$. We then obtain the linear predictors $\bA$ and $\bB$ using the EDMD approach presented in the EDMD section. 



We pick 50 new trajectories of length $t = 0.1 s$ generated using random control inputs from a normal distribution $[f_t, {\color{black}\bM}]^\top\sim \cN\bigl([0;0;0;0]^\top, \text{diag}([20;20;20;20])\bigr)$, which is {\color{black}different than the training set.} Then, the Koopman linear predictors are used to propagate the states starting the initial condition for 100 timesteps. The RMSE evaluated on the validation set is shown in Table \ref{tab:rmse_val}. It can be seen that the prediction errors are around 5$\%$, which indicates that the linear predictors $\bA$ and $\bB$ obtained through EDMD are a good representation of the original nonlinear system. A sample evaluation for tracking a random trajectory is shown in Fig. \ref{fig:EDMD}b

 \begin{table}[ht]
\caption{RMSE \% on validation set with $p=3$ observables}
\centering
\begin{tabular}{ c c c  } 
 \hline
 States                 & RMSE \% \\
\hline
 $\bp$                  & 3.30 $\pm$ 2.95 \\
 $\bv$                  & 5.57 $\pm$ 4.52\\
 $\mathbf{\Theta}$      & 5.59 $\pm$ 3.46\\
 $\bomega$                & 3.23 $\pm$ 2.15\\
 \hline
 average                & 4.52 $\pm$ 3.27\\
 \hline
\end{tabular}
\label{tab:rmse_val}
\end{table}
 
\subsection{SE(3) Koopman-MPC} \label{sec:MPC_results}
We design an MPC-based controller using the learned linear predictors $\bA$ and $\bB$ from EDMD as outlined in Section \ref{sec:MPC}. A random reference trajectory is generated with a sampling time of $t_s = 0.001$s with random control inputs from a normal distribution, $[f_t, {\color{black}\bM}]^\top\sim \cN\bigl([2;2;2;2]^\top, \text{diag}([30;30;30;30])\bigr)$, which is different from the training and validation distributions. We use a prediction horizon of $N_h = 10$ and a simulation time of $t = 1.2$s. The optimal control input at each timestep was obtained by solving the QP in a receding horizon fashion. Note that the lifted space with $p=3$ observables has a dimension of 51. New control inputs at the end of each optimization step are generated at the rate of 100 Hz.  The proposed framework was implemented using quadprog in MATLAB on a standard laptop with 8 GB RAM and an 8$^{th}$ generation intel core i5 CPU. {\color{black}Since the QP formulation can be easily implemented using C++, the MPC speed can be increased by several orders depending on the hardware capability}.

\begin{figure}[htbp]
     \centering
     \includegraphics[width = 0.75\linewidth]{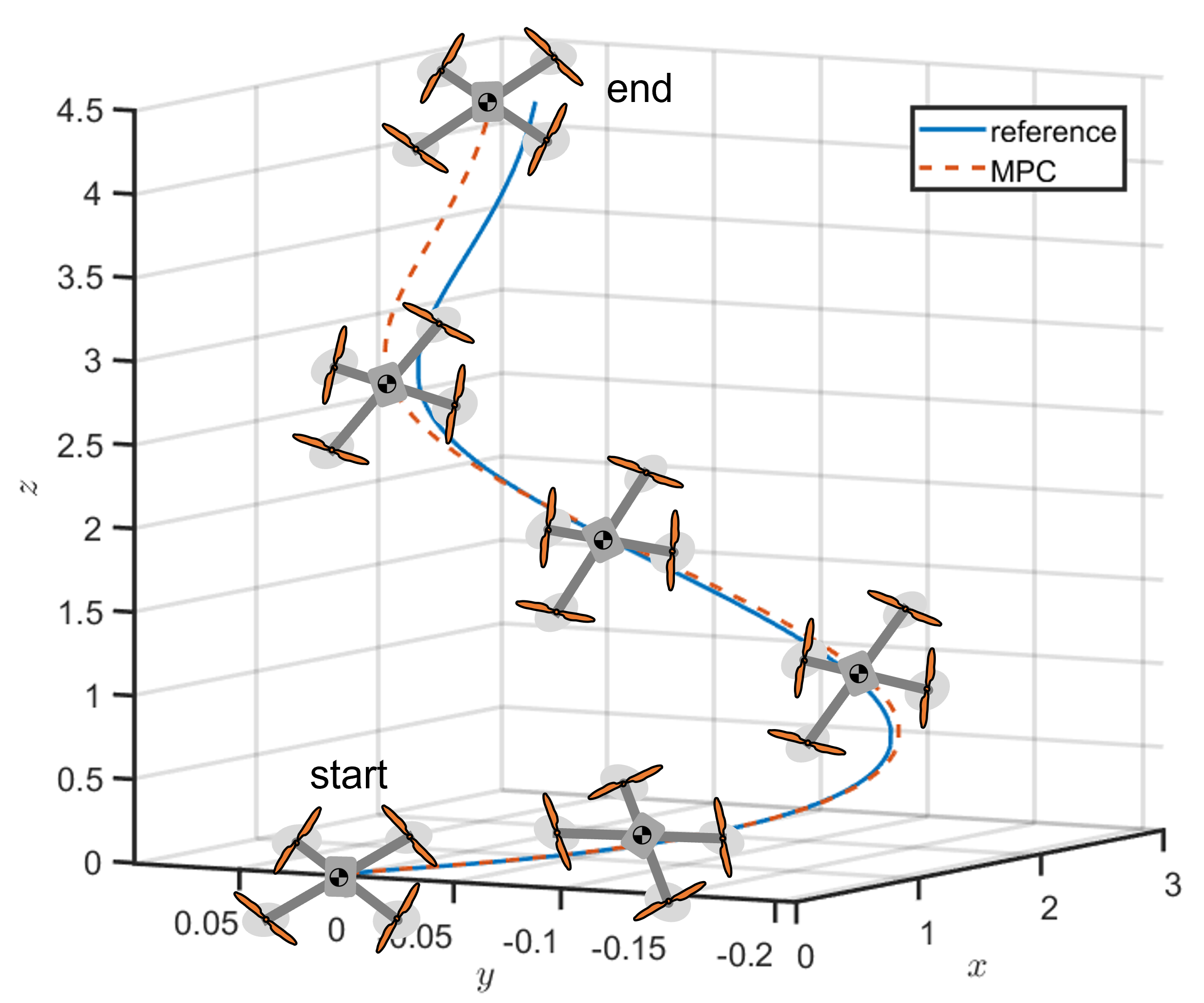}
     \caption{Trajectories obtained using SE(3) Koopman MPC can accurately track the reference trajectory.}
     \label{fig:MPC_traj}
 \end{figure}

\begin{figure}[htbp]
     \centering
     \includegraphics[width = 1\linewidth]{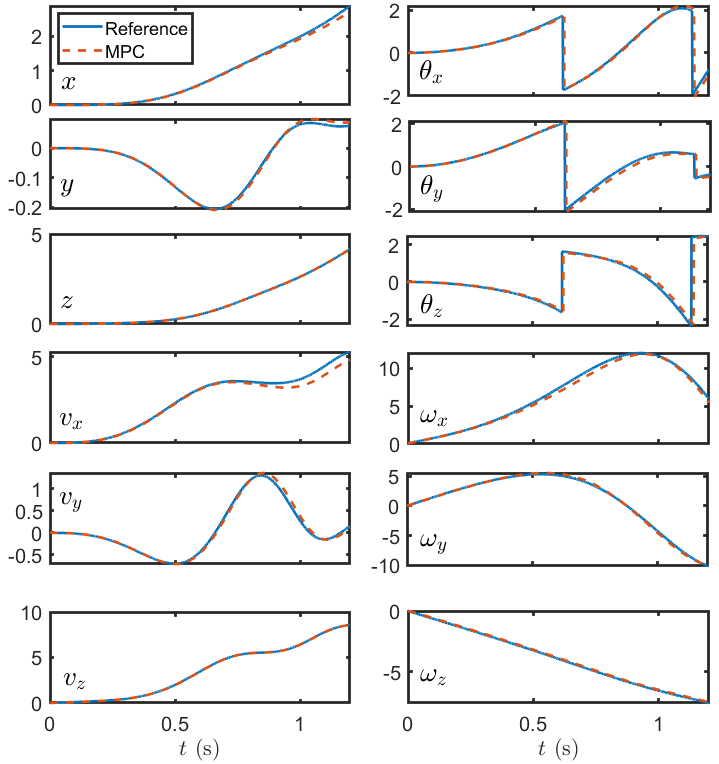}
     \caption{The state trajectories obtained from the MPC can accurately track the reference.}
     \label{fig:MPC_states}
 \end{figure}

\begin{figure}[htbp]
     \centering
     \includegraphics[width = 1\linewidth]{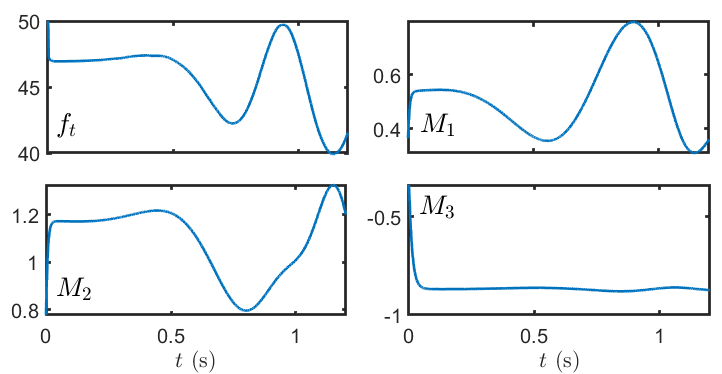}
     \caption{The corresponding control inputs obtained from the MPC stay within the control limits.}
     \label{fig:MPC_control}
 \end{figure}

Fig. \ref{fig:MPC_traj} shows that the MPC-based controller is able to track the given reference trajectory accurately while satisfying constraints. The quadrotor starts from the ground with zero velocities and can closely follow a random reference trajectory up to a height of 2 meters in under 1s.  In Fig. \ref{fig:MPC_states}, we can see the controlled trajectory (red dashed) closely follows the reference trajectory (black). Fig. \ref{fig:MPC_control} shows the corresponding control inputs $(f_t, \bM)$ for the linear and angular dynamics. The tracking is accurate up to $t = 1$s. However, the trajectories start to deviate from the reference after $t > 1 $s. The long-term prediction error can be improved using EDMD with online learning, which will be explored in future works.

\section{Conclusions} \label{conclusions}
In this work, we propose a set of physics-informed observables that can be used to obtain linear predictors for a nonlinear three-dimensional quadrotor system. The observable functions are constructed as products of rotation matrices and angular velocities. Augmenting this set of observables with the linear states ensures that the lifted system preserves the underlying SE(3) topology of the original nonlinear system. We show that the learned linear predictors using EDMD are able to predict arbitrary random trajectories with reasonable accuracy (with an average error of 4.5\%). Further, we use the learned linear predictors to design a controller to track references. To this extent, we develop a linear MPC in the lifted states. The proposed SE(3) MPC framework can be implemented in real-time at 100 Hz and is able to track arbitrary random reference trajectories with appropriate performance. Future works will incorporate EDMD with online learning in order to track trajectories in the presence of unknown disturbances such as wind.


\bibliography{ifacconf}   
\end{document}